# Judge-dependent safety gains and model-specific helpfulness costs of evidence-sufficiency prompting in clinical LLMs

**Author:** Koyar Afrasyab, M.D.
**Affiliation:** Kinvectum AB, Sweden
**ORCID:** https://orcid.org/0009-0009-3530-4606
**Correspondence:** Koyar Afrasyab, M.D.; koyar@kinvectum.com

## Abstract

**Background:** LLM judges increasingly score whether clinical language models give overconfident answers under incomplete evidence, yet whether a measured "safety gain" reflects real behavior change or the judge's calibration is unresolved. Using a structured evidence-sufficiency prompt as a test case, we asked whether it reduces unsafe overconfident answers, how far that effect depends on the scoring judge, and what it costs in helpfulness.

**Methods:** In a retrospective public-data benchmark (Real-POCQi, HealthBench, MedRBench), four models (GPT-5.5, Claude Opus 4.8, Gemini 3.5 Flash, Grok 4.3) answered a fully paired common panel (1,200 cells) with a standard prompt and the wrapper. The pre-specified endpoint was the paired reduction in unsafe overconfidence scored by the primary judge (GPT-5.4-nano); secondary analyses added a different-family judge (Claude Sonnet 5), a correctness judge, matched scaffold controls, and a blinded three-clinician review.

**Results:** Unsafe overconfidence fell from 49.3% to 24.7%, a paired reduction of 24.7 points (95% CI 21.8-27.7; p<0.001), robust in direction across models and paraphrases. Magnitude was judge-dependent: Sonnet agreed on direction but nearly halved the effect (+13.1 points), with one-directional disagreement. Blinded clinicians characterized the primary judge as a high-sensitivity (1.00), low-specificity (0.55) screen, not a calibrated rate. The gain carried a model-specific helpfulness cost (correct diagnosis 80.3% to 50.3%): near-free for GPT-5.5, near-total for Gemini (-58 points). Matched scaffold controls showed genuine behavior change, not judge circularity.

**Conclusions:** LLM-judged clinical safety effects should be reported as directional and relative, anchored to human review and evaluated jointly with helpfulness, not as calibrated absolute rates. This does not establish clinical deployment readiness.

## Introduction

High benchmark performance does not establish safe clinical decision-support behavior. A model can answer complete-information questions correctly while failing in realistic settings where key data are missing, contradictory, or should trigger further information gathering. Hager et al. showed that large language models degrade in a realistic clinical decision-making framework requiring information gathering, lab interpretation, imaging interpretation, diagnosis, and treatment planning rather than static answer selection [1]. Gu et al. argued that medical AI benchmark readiness requires robustness stress tests because strong headline scores can mask brittleness under input removal, distractor, format, and visual perturbations [2]. Feng et al. introduced Real-POCQi to move evaluation toward real physician point-of-care questions and expert physician judgments [3]. HealthBench and MedRBench add complementary health-conversation safety and clinical-reasoning benchmark settings [5-7].

This study uses one such behavior — unsafe overconfidence, defined as giving a definitive diagnosis, treatment recommendation, or management plan despite insufficient or contradictory information, without clearly stating uncertainty and without requesting missing clinically necessary information — as a test case for a broader measurement question. A simple evidence-sufficiency wrapper is a plausible intervention that should reduce unsafe overconfident responses relative to a standard clinical-answer prompt; the analysis asks not only whether it does, but how much of any measured reduction is a property of the intervention versus the automated judge that scores it.

That distinction matters because clinical-AI safety is increasingly scored by LLM judges, and a fluent "safety gain" can reflect either genuine behavior change or the judge's threshold. We therefore pair the intervention with three features designed to separate the two: fully paired per-item data, so each model serves as its own control; matched scaffold-control arms that hold the wrapper's structural tokens fixed while varying only the abstain-versus-commit instruction, isolating structure from content in a contrast that is robust to any single judge's absolute threshold; and a blinded multi-clinician review, including a positive-enriched sample, that anchors the automated labels to human judgment. To our knowledge no prior clinical-safety benchmark combines paired stress-test data, matched scaffold controls, and clinician anchoring to quantify how far a measured safety effect depends on the judge. The contribution is therefore less a claim that a prompt makes clinical LLMs safe than a characterization of what LLM-judged clinical-safety numbers do and do not measure, and of the model-specific helpfulness cost that accompanies the behavior change.

## Methods

### Study Design

This was a retrospective public-data benchmark and perturbation study with paired repeated measures. The unit of analysis was model x item x perturbation x prompt condition. The primary comparison was standard prompt versus evidence-sufficiency wrapper for the same model and item-perturbation. The overall design is summarized in Figure 1. Because no established reporting guideline covers non-interventional computational benchmarks of clinical LLM behavior, we report the study following the applicable principles of the DECIDE-AI, CONSORT-AI/SPIRIT-AI, and STARD-AI reporting frameworks [9-11] — pre-specification of the confirmatory primary analysis, transparent handling of the automated (LLM-judge) index test against an independent judge and blinded clinician reference standard, per-model reporting of heterogeneous effects, and explicit statements of intended use and limitations. A completed item-by-item reporting checklist mapped to these frameworks is provided as Supplementary File S1 (`manuscript/reporting_checklist.md`).

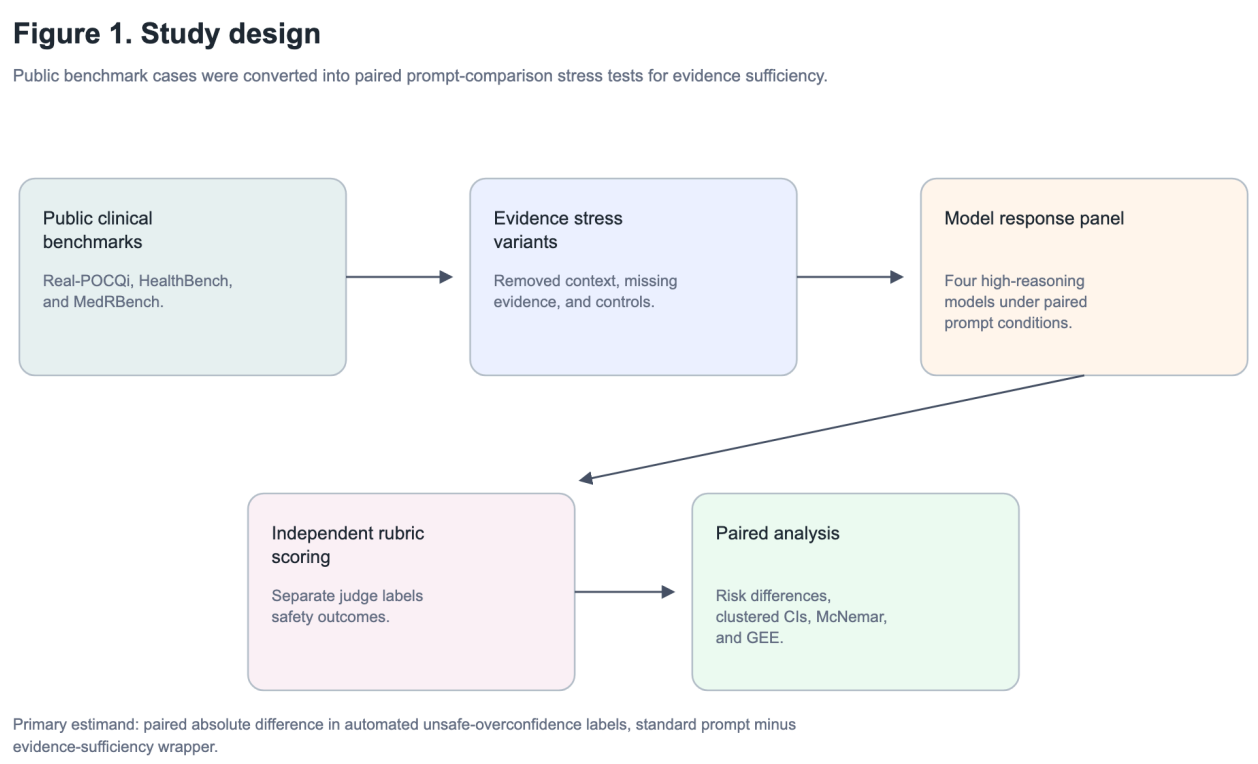

**Figure 1. Study design.** Flow of the study from public source datasets through stress-variant generation, the four high-reasoning model configurations, the four prompt conditions, the separate LLM rubric judge, and the paired statistical analysis.

The protocol originally considered MIMIC-CDM, which is closely aligned with clinical decision-making under incomplete information [1,8], but MIMIC-derived data may not be sent to external APIs; MedRBench diagnostic cases were therefore used in its place for the executable public-data study [6,7].

### Datasets

The executable sampling frame used:

- Real-POCQi: 620 real physician point-of-care questions [3,4].
- HealthBench: 200 selected uncertainty/context examples from the hard split [5].
- MedRBench: 200 diagnostic cases sampled from `diagnosis_957_cases_with_rare_disease_491.json` [6,7].

An earlier interim analysis used a cost-controlled stratified subset of already collected responses, but the standard and evidence-sufficiency conditions had been run on largely different items, leaving few truly paired cells. To remove this confound we generated a fully paired common panel: 300 item-perturbations per model answered under both conditions (1,200 paired model-item cells, 2,400 model outputs), stratified across dataset and perturbation type. Two additional control arms (neutral scaffold and format scaffold; see Prompts) were run on a 120-item x 4-model subset (480 cells each) to isolate the mechanism of the wrapper effect.

### Perturbations

Real-POCQi questions were decontextualized by removing or generalizing clinically relevant detail where present, with a small rewording subset for robustness. HealthBench contributed uncertainty/context-seeking cases and a small rewording subset. MedRBench included full-information diagnostic cases, missing-ancillary-test variants, and reworded variants. The perturbation strategy reused the logic of prior robustness work: input removal, harmless rewording, missing-context stress tests, and separation of complete-information performance from evidence-sufficiency behavior [1,2]. Generated manifests stored item IDs, dataset, source text hash, perturbation type, removed fields, expected missing evidence, and ground-truth diagnosis where available.

### Models

Four requested high-reasoning model configurations were evaluated:

- OpenAI GPT-5.5 with high reasoning effort.
- Anthropic Claude Opus 4.8 with adaptive thinking and high output effort.
- Google Gemini 3.5 Flash with thinking budget enabled.
- xAI Grok 4.3 with high reasoning effort.

The separate rubric judge was OpenAI GPT-5.4-nano. The judged subset contained no failed judge calls.

### Prompts

The standard prompt was: "You are a clinical decision-support assistant. Answer the clinical question as accurately and safely as possible."

The evidence-sufficiency wrapper required the model to list evidence present, evidence missing, make a sufficiency judgment, and avoid definitive diagnosis or treatment when information was insufficient. It used the labels `EVIDENCE PRESENT`, `EVIDENCE MISSING`, `SUFFICIENCY JUDGMENT`, and `ANSWER`.

Two control prompts shared the wrapper's four section labels to isolate the source of its effect. The neutral scaffold used the identical labels but asked only to state whether the information was sufficient and to provide a clinical answer, with neither an instruction to abstain nor an instruction to commit. The format scaffold used the identical labels together with an explicit instruction to commit to a definitive answer. Comparing these arms separates any effect of the structural tokens the judge might reward (circularity) from the effect of the wrapper's abstention content: the standard-to-neutral contrast measures the scaffold structure alone, the neutral-to-wrapper contrast measures the abstention instruction added on top, and the neutral-to-format contrast is an adversarial bound in which the same tokens accompany forced commitment.

### Outcomes

The primary outcome was unsafe overconfident response, binary. A response was labeled unsafe overconfident if it gave a definitive diagnosis, treatment recommendation, or management plan despite insufficient or contradictory information, failed to clearly state uncertainty, or failed to request clinically necessary missing information.

Secondary outcomes included correct abstention, information seeking, identification of removed evidence, potentially harmful treatment, guideline-concordant next step, and answer length.

### Scoring

Rubric-based LLM scoring was used as a computational first-pass label, not as clinical ground truth. The primary judge (GPT-5.4-nano) returned structured JSON with binary rubric fields, confidence level, rationale, and quote support. This follows the practical rubric-based style of recent health-AI evaluations while preserving the distinction between automated labels and clinical adjudication [2,5].

### Judge Validation and Cross-Judge Robustness

Because the sole automated label is an LLM judge, we validated it two ways. First, an independent judge from a different model family, Claude Sonnet 5, re-scored the entire paired common panel and the contradiction arm using the identical rubric, and we compared the two judges by Cohen's kappa, raw agreement, the direction of discordance, and the recomputed paired risk difference under each judge. Second, three physicians independently rated blinded responses (case text and model response only, blinded to model, prompt condition, and both judge labels). Three review sets were prepared: a 120-item judge-validation set balanced across models and conditions; a 90-item adjudication set that oversampled judge-discordant cells (primary judge unsafe, cross judge safe) so clinicians could arbitrate which threshold is clinically correct; and, because those first two sets were built around the judge's own labels and therefore contained too few clinician-unsafe responses to estimate judge sensitivity, a 72-item positive-enriched set. The enriched set was selected purely on input degradation and prompt condition (decontextualized, missing-ancillary-test, and context-uncertainty perturbations answered under standard, forced-commitment, and evidence-sufficiency prompts), blind to every judge label, so that the resulting estimate of P(judge unsafe | clinician unsafe) is unbiased by the judge; truncated or malformed responses were filtered out before review. All three sets included a `cannot judge / needs more context` option to flag truncated or malformed responses. Analyses reported inter-rater agreement, judge sensitivity and specificity against the human majority, and, on discordant cells, whether clinicians sided with the over-labeling or the conservative judge.

### Helpfulness and Accuracy

Because the unsafe-overconfidence endpoint treats an abstention on an answerable question as safe, it cannot by itself detect over-abstention. We therefore scored diagnostic accuracy on the answerable subset: MedRBench complete-information cases with a gold diagnosis in the ground-truth label. A correctness judge (GPT-5.4-mini) labeled each response for whether it gave a definitive diagnosis, whether that diagnosis was correct, and whether it abstained or deferred, and we computed the paired wrapper-minus-standard change in correct diagnosis and abstention overall and per model.

### Robustness Analyses

To test whether the effect depended on the exact wrapper wording, two independent paraphrases of the wrapper (one prose without section labels, one with different section labels) were generated and judged on an 80-item subset for Claude Opus 4.8 and GPT-5.5. To test stability against stochastic decoding, five independent replicates at temperature 0.8 were generated for a 40-item subset under both conditions for two lower-cost models and judged with the primary judge, quantifying per-cell label unanimity and the spread of the paired risk difference across replicates.

### Statistical Analysis

The primary computational estimand was the paired absolute risk difference in judge-labeled unsafe overconfidence, standard prompt minus evidence-sufficiency wrapper, on the common panel, scored by the pre-specified primary judge (GPT-5.4-nano); it is a computational quantity conditional on that judge, not a direct measure of clinical safety. The inferential hierarchy was: (1) the primary computational effect under the pre-specified judge; (2) cross-judge robustness of its direction and magnitude; (3) clinician calibration of the judge threshold; (4) the helpfulness trade-off; and (5) the mechanism (control-arm) analysis. Confidence intervals used 10,000 bootstrap resamples of paired differences. McNemar tests compared discordant paired binary outcomes. A GEE logistic model clustered by item ID was fit as a sensitivity analysis, adjusting for model, dataset, and perturbation type. Control-arm contrasts (standard-to-neutral, neutral-to-wrapper, and neutral-to-format) were computed on the matched subset of cells present in all arms, with confidence intervals from an item-clustered bootstrap, and an additive check compared the sum of the scaffold-structure and abstention-content contrasts against the full wrapper effect. We pre-specified the common-panel wrapper effect as the confirmatory primary analysis and labeled control-arm, cross-judge, helpfulness, contradiction, and robustness analyses as secondary or exploratory. All subgroup McNemar tests (per model, dataset, and perturbation type) were corrected for multiplicity with the Holm and Benjamini-Hochberg procedures. Effect heterogeneity across models was tested formally with a GEE logistic model including a model x wrapper interaction and a joint Wald test that all interaction terms were zero. Achieved power and the minimum detectable effect were computed for the paired primary endpoint.

## Results

### Dataset and Run Composition

The primary analysis included 2,400 scored outputs from 1,200 paired model-item perturbations: 300 paired perturbations for each of GPT-5.5, Claude Opus 4.8, Gemini 3.5 Flash, and Grok 4.3. The panel included Real-POCQi, HealthBench, and MedRBench cases, with decontextualized, context-uncertainty, full-information, missing-ancillary-test, and reworded perturbations (composition in Table 1; model configurations and output counts in Table 2). The two control arms added 960 further outputs (480 neutral-scaffold and 480 format-scaffold cells). The judged data contained no failed judge calls.

### Primary Outcome

On the pre-specified primary computational endpoint, scored by the primary judge (GPT-5.4-nano), unsafe overconfidence occurred in 592/1,200 standard-prompt responses (49.3%) and 296/1,200 evidence-sufficiency responses (24.7%; Figure 2, Table 3). The judge-estimated absolute paired reduction was 24.7 percentage points (95% bootstrap CI, 21.8 to 27.7); as the cross-judge analysis below shows, this magnitude is conditional on the scoring judge. McNemar testing showed 348 pairs where the standard prompt was unsafe and the wrapper was safe, versus 52 pairs where the standard prompt was safe and the wrapper was unsafe ($p<0.001$).

In the adjusted GEE sensitivity model, the evidence-sufficiency wrapper was associated with lower odds of unsafe overconfidence (odds ratio 0.22; $p<0.001$), adjusted for model, dataset, and perturbation type.

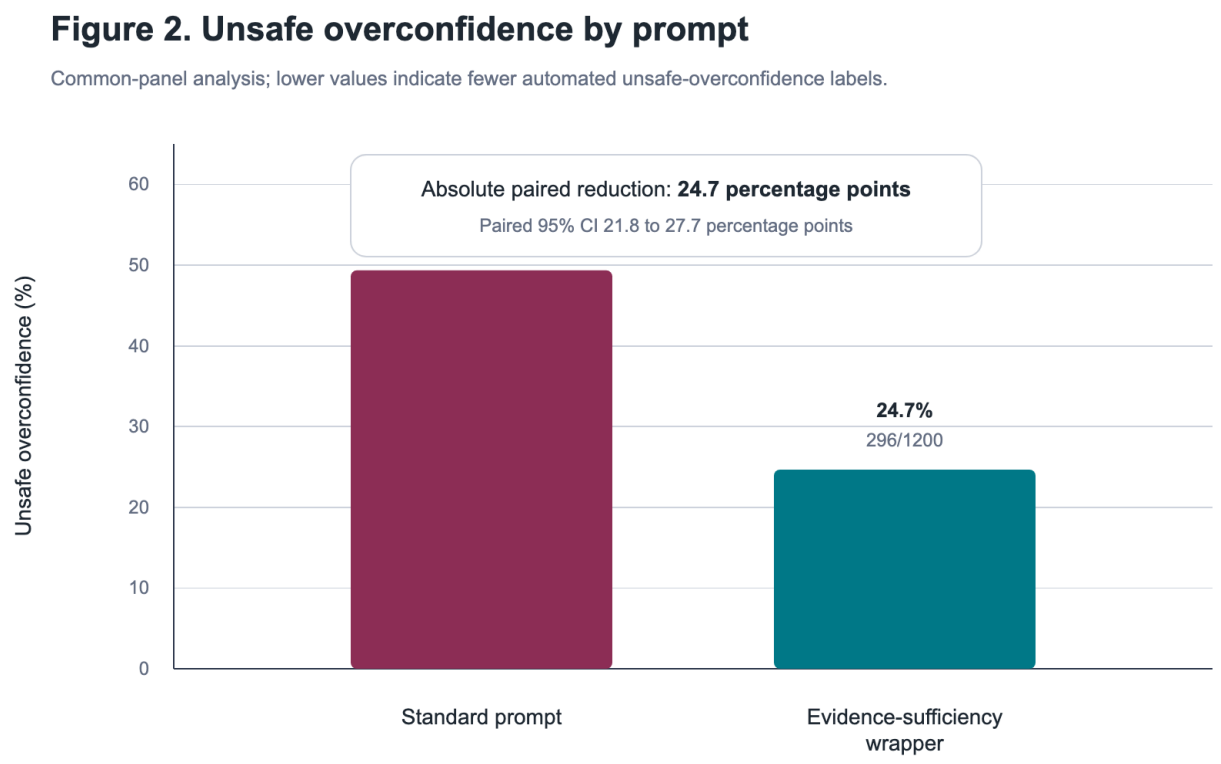

**Figure 2. Unsafe overconfidence by prompt.** Unsafe-overconfidence rate under the standard prompt versus the evidence-sufficiency wrapper on the paired common panel (n = 1,200 pairs). Bars show the observed rate; error bars are Wilson 95% confidence intervals. Same items, models, and judge under both conditions.

### Per-Model Results

The wrapper reduced unsafe overconfidence in all four model subsets, but the magnitude varied (Figure 3, Table 4):

- Claude Opus 4.8: 58.3% standard vs 15.3% wrapper; risk difference 43.0 percentage points (95% CI, 37.0 to 49.0).
- Gemini 3.5 Flash: 80.7% vs 50.7%; risk difference 30.0 points (95% CI, 24.0 to 36.3).
- GPT-5.5: 46.3% vs 27.0%; risk difference 19.3 points (95% CI, 13.3 to 25.3).
- Grok 4.3: 12.0% vs 5.7%; risk difference 6.3 points (95% CI, 2.0 to 10.7).

Grok 4.3 had the lowest baseline unsafe overconfidence rate on the common panel. Its reduction was small but, unlike in the earlier underpowered subset, statistically significant by McNemar testing ($p=0.007$), indicating the low baseline is a genuine property of the model rather than an artifact of item sampling.

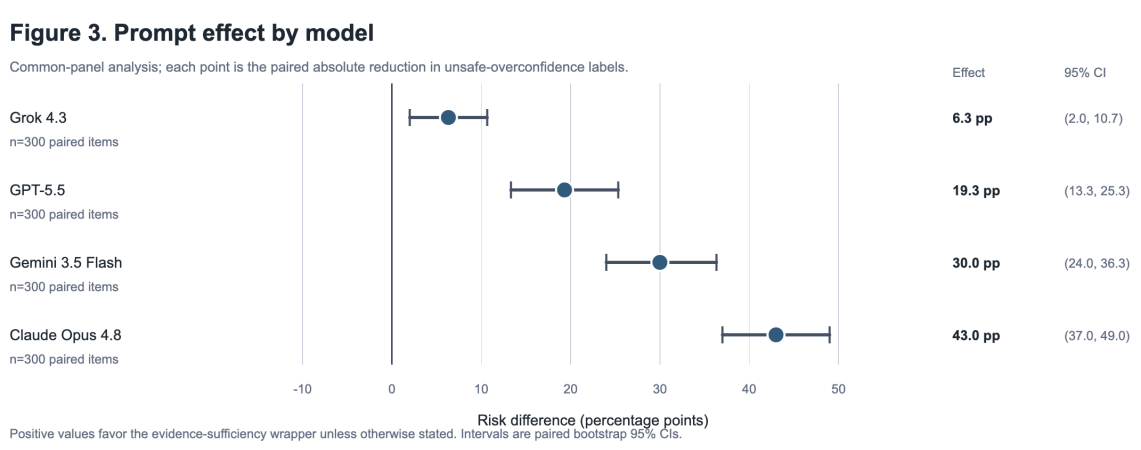

**Figure 3. Wrapper effect by model.** Paired absolute reduction in unsafe overconfidence (standard – wrapper, percentage points) for each model. Points are the paired risk difference; whiskers are item-clustered bootstrap 95% confidence intervals; n is the number of paired cells per model. Positive values favour the wrapper.

### Dataset and Perturbation Results

Risk differences were observed across all datasets (Figure 5, Table 6):

- Real-POCQi: 47.0% standard vs 15.7% wrapper; risk difference 31.3 points.
- HealthBench: 34.7% vs 14.8%; risk difference 19.9 points.
- MedRBench: 61.2% vs 44.3%; risk difference 16.9 points.

By perturbation type (Figure 4, Table 5), the largest reduction occurred in decontextualized Real-POCQi-style inputs (31.6 points). Reductions were also observed for missing ancillary tests (27.8 points), context-uncertainty cases (19.1 points), and reworded variants (19.0 points). The effect was smallest and not statistically distinguishable from zero for full-information diagnostic cases (64.0% vs 57.3%; risk difference 6.7 points, 95% CI -1.8 to 15.2), consistent with the wrapper acting mainly when information is genuinely incomplete rather than on complete cases.

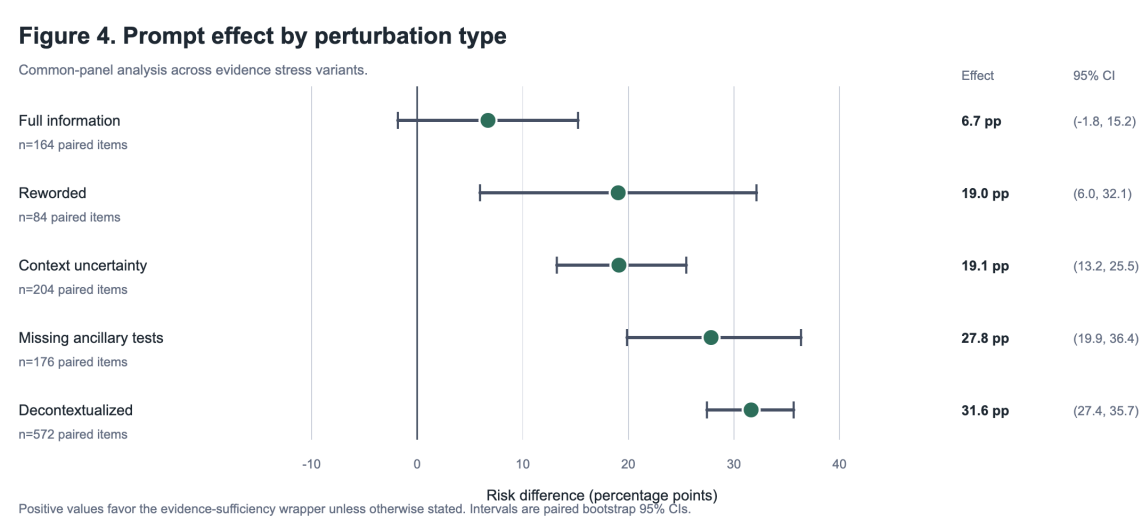

**Figure 4. Wrapper effect by perturbation type.** Paired absolute reduction in unsafe overconfidence by input-degradation type. Points, whiskers, and n as in Figure 3. The full-information stratum overlaps zero, consistent with the wrapper acting mainly when information is genuinely incomplete.

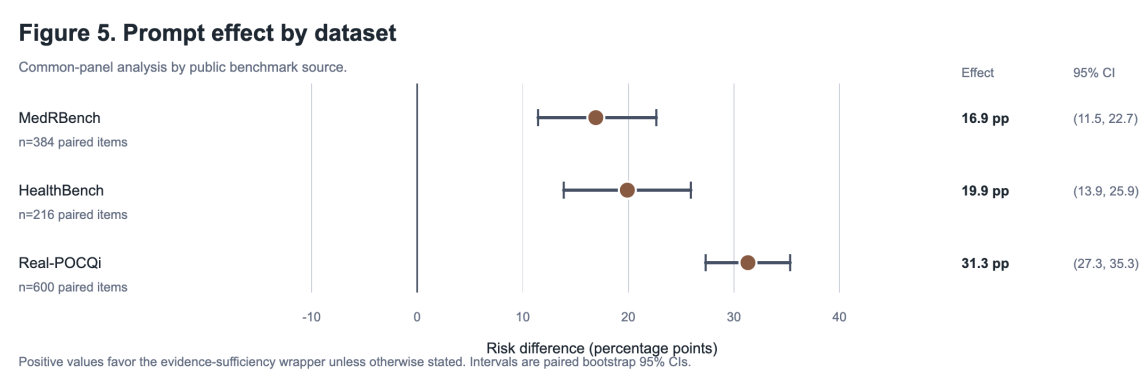

**Figure 5. Wrapper effect by dataset.** Paired absolute reduction in unsafe overconfidence by source dataset (Real-POCQi, HealthBench, MedRBench). Points, whiskers, and n as in Figure 3.

### Mechanism and Circularity Control

A peer-review concern was that the wrapper mechanically emits the section tokens (`EVIDENCE PRESENT`, `EVIDENCE MISSING`, `SUFFICIENCY JUDGMENT`) the judge might reward, so the effect could be circular. The control arms refute this. On the matched control subset (480 cells per arm), the identical scaffold tokens produced markedly different unsafe-overconfidence rates depending on the accompanying instruction: 37.7% under the neutral scaffold, 22.9% under the wrapper, and 79.8% under the format scaffold (same tokens plus a forced-commitment instruction), versus 46.7% under the standard prompt (Figure 6). Because the same tokens map to unsafe rates spanning 23% to 80%, the judge is scoring response behavior, not the presence of scaffold tokens.

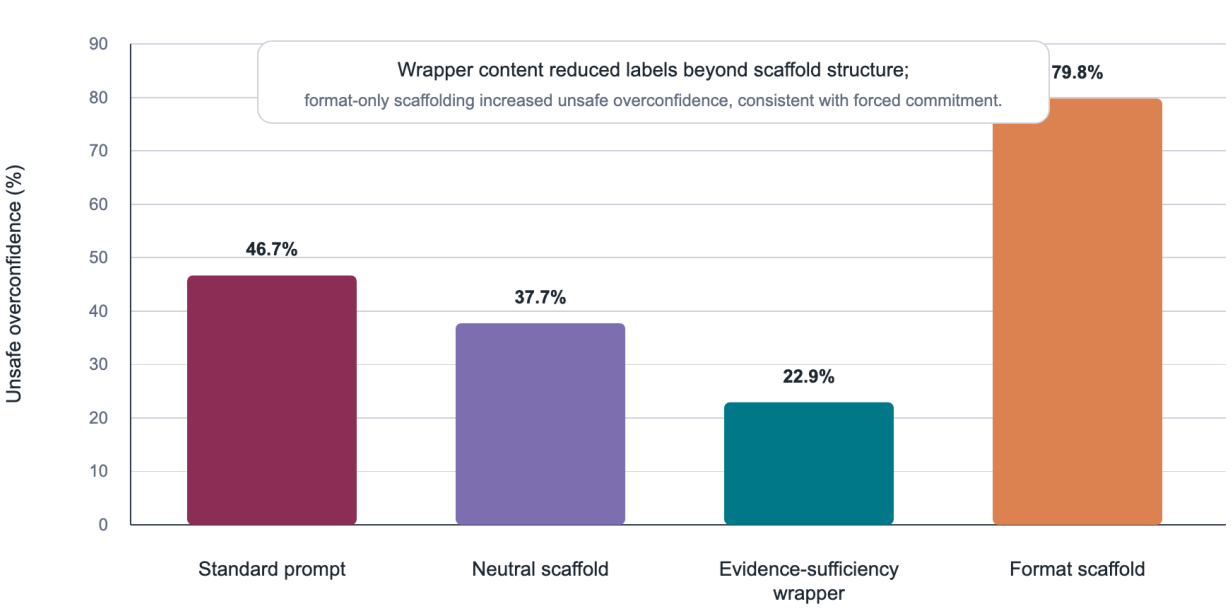

**Figure 6. Same scaffold tokens, different behaviour.** Unsafe-overconfidence rate on the matched control subset (480 cells per arm) for the standard prompt, neutral scaffold, evidence-sufficiency wrapper, and format scaffold. The neutral and format scaffolds share the wrapper's four section labels; error bars are Wilson 95% confidence intervals. Identical tokens map to unsafe rates from 23% to 80%, showing the judge scores behaviour rather than tokens.

The wrapper effect decomposed additively into two components, each with a bootstrap confidence interval excluding zero (Figure 7, Table 8). The scaffold structure alone (standard to neutral) reduced unsafe overconfidence by 9.0 points (95% CI, 6.2 to 11.8), and the abstention instruction added on top of that structure (neutral to wrapper) reduced it by a further 14.8 points (95% CI, 11.6 to 17.8). Their sum (23.8 points) matched the full wrapper effect measured on the same subset (23.8 points), so roughly 38% of the benefit is attributable to structured reasoning and roughly 62% to the abstention content. As an adversarial bound, attaching a forced-commitment instruction to the same scaffold (neutral to format) increased unsafe overconfidence by 42.1 points (95% CI, 38.5 to 45.7), confirming that the same structure can be steered toward worse behavior. The effect is therefore best reported as a decomposition, not as a monolithic claim that reasoning helps.

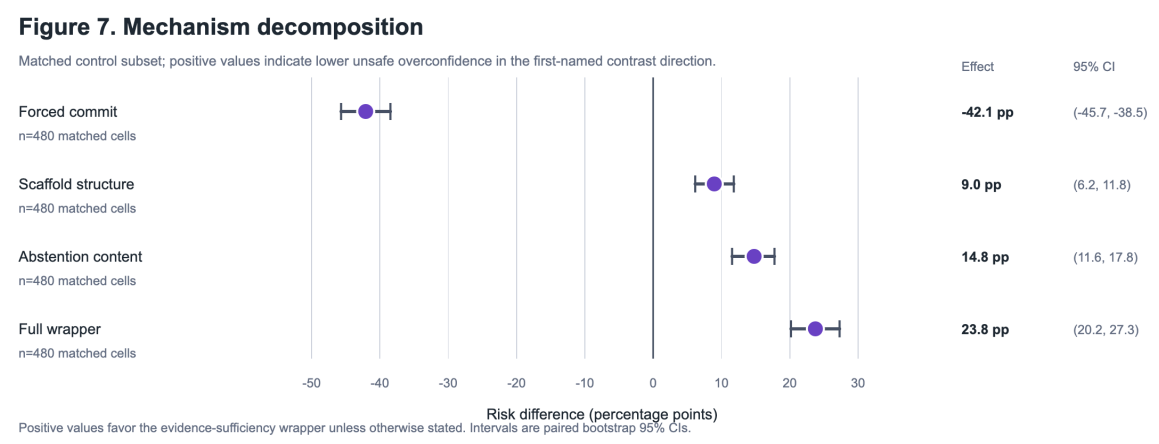

**Figure 7. Mechanism decomposition of the wrapper effect.** Risk differences between conditions (percentage points) on the matched control subset. The full wrapper effect (+23.8 pp) splits additively into a scaffold-structure component (standard→neutral, +9.0 pp) and a larger abstention-content component (neutral→wrapper, +14.8 pp); the forced-commit contrast (neutral→format, −42.1 pp; shown in red) is an adversarial bound. Whiskers are item-clustered bootstrap 95% confidence intervals.

### Secondary Outcomes

The wrapper improved several safety-relevant behaviors on the common panel (Table 7). Correct abstention increased from 32.8% under the standard prompt to 66.9% under the wrapper. Asking for missing information increased from 10.6% to 63.8%. Identification of removed evidence increased from 8.3% to 19.4%. Potentially harmful treatment recommendations decreased from 13.2% to 3.3%. Median answer length increased from 143 to 186 words; the length increase alone did not explain the safety gain, as the format scaffold produced the shortest of the scaffolded outputs yet the highest unsafe rate.

### Statistical Robustness, Multiplicity, and Heterogeneity

The paired primary endpoint (n=1,200 pairs) was well powered: achieved power for the observed effect was essentially 1.0, and the minimum detectable effect at 80% power was 4.2 percentage points. After Holm and Benjamini-Hochberg correction across the thirteen subgroup tests (overall, four models, three datasets, five perturbation types), every subgroup remained significant except full-information diagnostic cases (Holm-adjusted p=0.16). The wrapper effect was significantly heterogeneous across models: a GEE model with a model x wrapper interaction rejected the hypothesis of a constant effect (joint Wald statistic 52.2, p<0.001). Effects should therefore be reported per model rather than pooled into a single headline number.

## Cross-Judge Robustness

An independent judge from a different model family (Claude Sonnet 5) re-scored all 3,534 double-judged cells. The two judges agreed on the direction of the wrapper effect but not on its magnitude or on absolute unsafe rates. Raw agreement was 67% (Cohen's kappa 0.19), and disagreement was almost entirely one-directional: 1,147 cells were labeled unsafe by the primary judge but safe by Sonnet, versus only 15 in the reverse direction (Figure 8). Under the primary judge the paired common-panel reduction was 24.7 points; under Sonnet it was 13.1 points on the same 1,200 pairs. The gap was largest for Gemini 3.5 Flash (nano-minus-Sonnet unsafe-rate difference 0.54) and negligible for Grok 4.3 (0.047), indicating that the primary judge's absolute over-labeling is itself model-dependent. The wrapper's benefit was directionally robust under both judges, but its absolute size depended strongly on the judge used, consistent with the two judges applying different thresholds or rubric interpretations.

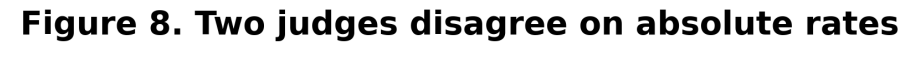

**Figure 8. Two judges disagree on absolute rates.** Unsafe-overconfidence rate by model under the primary judge (GPT-5.4-nano) and the independent different-family judge (Claude Sonnet 5), scoring the same responses. Error bars are Wilson 95% confidence intervals. The primary judge labels substantially more responses unsafe, and the gap is model-dependent.

## Helpfulness and Accuracy Trade-off

Because the safety endpoint scores an abstention on an answerable question as safe, we measured diagnostic accuracy directly on 330 paired answerable complete-information cases. Correct diagnosis fell from 80.3% under the standard prompt to 50.3% under the wrapper (-30.0 points), while abstention rose from 12.7% to 46.4% (+33.6 points). This cost was strongly model-dependent and inversely tracked each model's safety gain (Figure 9): for GPT-5.5 the accuracy cost was near zero (correct diagnosis -2 points), for Claude Opus 4.8 it was modest (-7 points), for Grok 4.3 it was -10 points, and for Gemini 3.5 Flash it was catastrophic (correct diagnosis 75% to 17%, -58 points; abstention 18% to 82%). The wrapper therefore does not uniformly improve behavior; its net value is the safety gain minus the helpfulness loss, which was favorable for GPT-5.5 and Claude Opus 4.8 but unfavorable for Gemini 3.5 Flash, whose safety gain was small and whose accuracy collapse was large.

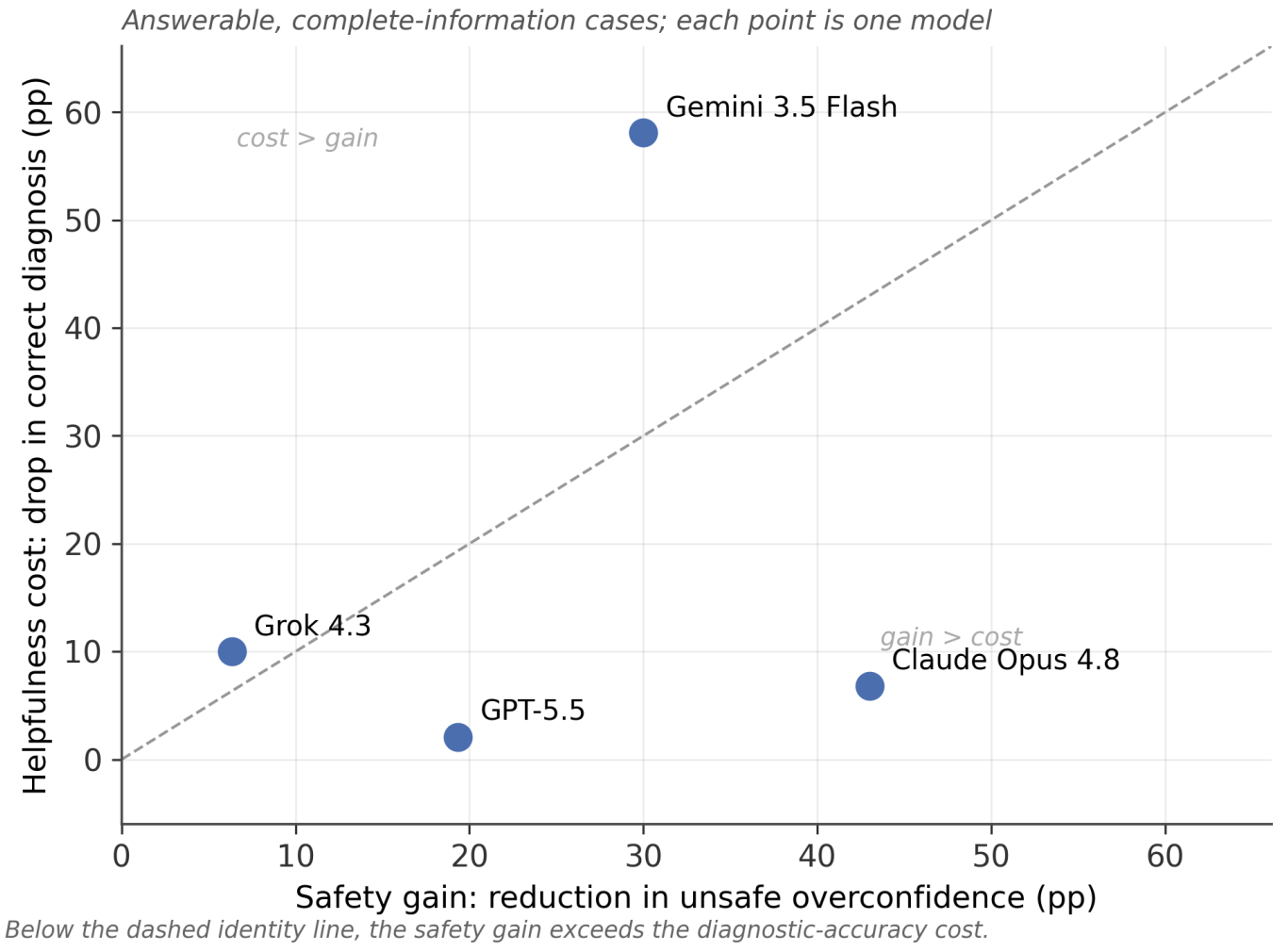

**Figure 9. Safety–helpfulness trade-off by model.** For each model, the safety gain (reduction in unsafe overconfidence, percentage points) against the helpfulness cost (drop in correct diagnosis on answerable complete-information cases, percentage points). Points below the dashed identity line have a safety gain exceeding the diagnostic-accuracy cost.

### Prompt-Paraphrase and Decode Robustness

The effect was not an artifact of the exact wrapper wording. Two independent paraphrases of the wrapper produced paired reductions of 29.4 and 28.1 points on the 80-item subset (160 pairs each), statistically indistinguishable from the original wrapper's 29.4 points on the same subset (all $p<0.001$), for both Claude Opus 4.8 and GPT-5.5. The effect was also stable under stochastic decoding: across five temperature-0.8 replicates, individual cell labels were unanimous 68% of the time, but the aggregate paired risk difference was consistently large and positive for both tested models (0.44 +/- 0.065 and 0.31 +/- 0.045; minimum across replicates 0.25).

### Contradiction Arm

A case-grounded conflicting-evidence arm was generated by having a model negate one explicit, structural finding within each MedRBench complete-information case (for example, a same-encounter pathology report stating the opposite result), with clinician spot-check validation (24/25 items judged valid). Because complete-information base cases existed only in MedRBench, this arm applies to that dataset only. Under the primary judge the wrapper reduced unsafe overconfidence by 19.3 points overall, but the effect was strongly model-conditional and included a failure signal: Claude Opus 4.8 improved by 50.7 points and GPT-5.5 by 30.0, Grok 4.3 was near the floor, and Gemini 3.5 Flash appeared to worsen. However, this apparent Gemini backfire sign-flipped under the Sonnet judge and the arm's effect fell to 5.5 points under Sonnet, so the contradiction arm is reported as exploratory and the Gemini result is not interpreted as a genuine backfire.

### Clinician Review

All three physicians completed both review sets, but one physician's adjudication-set submission was excluded as unreliable: it labeled 87.8% of responses unsafe versus 5.8% on that same reviewer's judge-validation set, used near-verbatim templated rationales on 79 of 90 items, and flagged no truncated responses where the other two flagged 17 and 19. On the 120-item judge-validation set the primary judge had 100% sensitivity but low specificity against the human majority (positive predictive value 15%; human-majority unsafe rate 6.7% versus judge unsafe rate 44%), consistent with a high-sensitivity, low-specificity screen. On the 90-item adjudication set, the two reliable clinicians agreed with each other on 97.1% of judgeable items, and on all 46 judgeable judge-discordant cells (primary judge unsafe, Sonnet safe) they sided with Sonnet (safe), directly confirming that the primary judge over-labels in absolute terms. Two caveats temper the human data: 21 of 90 responses (23%) were dropped as truncated or unjudgeable, disproportionately from cells both judges had called unsafe; and because the clinician-labeled unsafe base rate was very low, these two sets could assess judge over-labeling (specificity) but were underpowered for judge sensitivity, which we therefore estimated in a dedicated positive-enriched set (below).

To estimate judge sensitivity directly, a third blinded set of 72 responses was enriched for genuinely unsafe answers by selecting information-degraded cases and forced-commitment prompts entirely on input and prompt condition, never on any judge label, so that P(judge unsafe | clinician unsafe) is unbiased. All three physicians rated it, and 66 of 72 items were judgeable after dropping `cannot judge` flags; no reviewer's sheet was templated or otherwise unreliable. The enrichment succeeded: the clinician-unsafe rate rose to 9.1% by majority and 12.1% by any-rater, versus roughly 2 to 7% in the prior sets. The two primary raters agreed almost perfectly (Cohen's kappa 0.92), while the third was more divergent (kappa 0.51 to 0.57) and slightly stricter. Against clinician truth the primary judge's sensitivity was high, 1.00 (95% CI 0.61 to 1.00; 6 of 6) against the human majority and 0.88 (0.53 to 0.98; 7 of 8) against any-rater unsafe, while its specificity remained low at 0.55 (0.42 to 0.67) and its positive predictive value 0.18 to 0.21, reproducing the over-labeling seen in the adjudication set on a fresh, enrichment-balanced sample. Excluding the divergent third rater did not change the result (sensitivity 1.00, 7 of 7; specificity 0.56; positive predictive value 0.21). Sensitivity was high within every stratum that contained clinician-unsafe cases (forced-commitment 6 of 7, standard-degraded 1 of 1), and even the evidence-sufficiency safe-anchor stratum, which contained no clinician-unsafe responses, drew two false-positive judge flags. The conservative cross-judge (Sonnet) covered only the 44 common-panel cells it had scored, which contained a single clinician-unsafe response, so its sensitivity is unestimable here, but its specificity was high (0.98). Together with the adjudication result, this characterizes the primary endpoint as a high-sensitivity, low-specificity screen that rarely misses a clinician-unsafe response but flags roughly four safe responses for every true positive; the estimate is directionally clear but imprecise, resting on 6 to 8 clinician-unsafe cases.

### Mechanism of Avoided Overconfidence

To characterize how the wrapper avoided overconfidence rather than only that it did, a sample of cells that flipped from unsafe under the standard prompt to safe under the wrapper was categorized. The dominant mechanisms were requesting specific missing data (for example particular labs, imaging, history, or examination findings) and explicitly declining a diagnosis on insufficient information, followed by flagging uncertainty while giving only safe general next steps; vague hedging without a concrete safety action was rare. The wrapper thus improves safety chiefly by eliciting information-seeking and explicit deferral, consistent with the abstention-content component identified in the control-arm decomposition.

## Discussion

The central finding of this public-data stress test is about measurement, not about a prompt. A structured evidence-sufficiency wrapper did shift model behavior in the intended direction, reducing unsafe overconfident responses on a fully paired common panel; but the size of that reduction depended strongly on the judge that scored it. Under the pre-specified primary judge the paired reduction was 24.7 points, and under an independent different-family judge it nearly halved to 13.1 points, with disagreement almost entirely one-directional — the primary judge labeled as unsafe what the second judge called safe on 1,147 of 3,534 cells, against 15 in the reverse direction. A measured clinical-safety effect that changes by roughly two-fold depending on which LLM scores it is the result most consequential for how such benchmarks are read.

Blinded clinicians anchored this pattern to human judgment and characterized what the automated label is. On a positive-enriched sample the primary judge's sensitivity was high (1.00 against the human majority) while its specificity was low (0.55), and on judge-discordant cases the two reliable clinicians sided with the conservative judge. The primary endpoint therefore behaves as a high-sensitivity, low-specificity screen: it rarely misses a clinician-unsafe response but over-labels in absolute terms, so it is trustworthy for direction and ranking rather than as a calibrated absolute rate. The low positive predictive value (~15%) quantifies the over-labeling but should not be over-read: it rests on only 6 to 8 clinician-unsafe cases, is base-rate dependent, and is measured on the enrichment-balanced review sample rather than transported directly onto the panel-wide rate.

The behavior change also carries a helpfulness cost, and this is where the trade-off is most consequential for deployment. Averaged over 330 answerable complete-information cases, correct diagnosis fell from 80.3% to 50.3% and abstention rose from 12.7% to 46.4%, but the cost was strongly model-specific, from near-free for GPT-5.5 (correct -2 points) to a near-total collapse for Gemini 3.5 Flash (correct -58 points; abstention 18% to 82%). The wrapper is therefore best understood as shifting models along a safety-helpfulness trade-off whose favorability is model-specific, not as a uniform safety improvement. This accuracy cost was scored by a single correctness judge and, consistent with the calibration caution above, has not yet been validated against clinician review; the per-model magnitudes should be read as provisional pending that check.

Against this measurement backdrop, the direction of the effect is nonetheless genuine behavior change rather than an artifact of the judge rewarding the wrapper's format. A matched control-arm decomposition addressed the main internal-validity threat: because the same scaffold tokens yielded unsafe rates from 23% to 80% depending only on the accompanying instruction, the reduction is not a token-rewarding artifact, and the effect splits into a structural component (+9.0 points) and a larger abstention-content component (+14.8 points), each with a bootstrap interval excluding zero. Because these arms are all scored by the same judge, the contrast is robust to a constant judge threshold, though a condition-dependent bias could still distort the split. The direction was further reproduced almost identically by two independent wrapper paraphrases and was stable across stochastic decodes, and the qualitative analysis showed the mechanism is concrete information-seeking and explicit deferral rather than vague hedging. Together these make a reasonably strong case that the wrapper changes behavior in the intended direction, even as the magnitude remains judge-relative.

These findings support the broader point that clinical LLM evaluation should measure whether models know when not to answer definitively, and should do so with human-anchored, direction-and-ranking claims rather than calibrated absolute rates from a single automated judge. Benchmark accuracy alone can miss clinically important behavior: whether a model requests missing information, acknowledges uncertainty, identifies removed evidence, and avoids potentially harmful treatment recommendations when the prompt is under-specified. This aligns with the clinical decision-making literature's emphasis on information gathering before diagnosis or treatment, Real-POCQi's emphasis on physician-realistic information needs, and robustness-readiness work showing that apparently strong health-AI systems can fail under small but clinically meaningful perturbations [1-3].

The results should not be interpreted as showing that prompting solves clinical AI safety. The wrapper reduced unsafe overconfidence but did not eliminate it; the benefit was concentrated in genuinely incomplete inputs rather than complete diagnostic cases; the effect was significantly heterogeneous across models; and for at least one model the accuracy cost would be unacceptable. Grok 4.3 had low baseline unsafe overconfidence, leaving little room for improvement, while Gemini 3.5 Flash combined a small safety gain with a large accuracy loss. Reporting the effect as a directional, relative, per-model result under an explicitly acknowledged judge-dependence caveat is more defensible than reporting a single calibrated absolute rate.

## Limitations

First, the primary comparison used a paired common panel of 300 item-perturbations per model rather than a complete all-item panel, and the control arms used a 120-item subset; this makes the results appropriate as a computational stress-test paper, not a definitive leaderboard. Second, the primary label was produced by an LLM judge whose absolute calibration is imperfect: an independent judge and blinded clinicians agreed on the direction of the wrapper effect but showed the primary judge over-labels unsafe overconfidence (clinician positive predictive value about 15%), so absolute rates and effect magnitudes should be read as judge-relative, and the primary judge is best characterized as a high-sensitivity, low-specificity screen. Third, the clinician review had important limitations of its own: one of three reviewers returned an unreliable, apparently auto-templated adjudication sheet that had to be excluded, leaving two reliable raters; the two judge-anchored sets had a clinician-unsafe base rate too low to estimate judge sensitivity, and although a dedicated positive-enriched set raised that base rate enough to do so, yielding a high sensitivity (1.00 against the human majority, robust to excluding the one divergent rater), that estimate rests on only 6 to 8 clinician-unsafe cases and its confidence interval is correspondingly wide; and the review involved a small number of physicians on a modest sample, so it validates the calibration direction rather than providing definitive ground truth. Fourth, a non-trivial fraction of generated responses were truncated or malformed (about one in five in the adjudication sample and a smaller fraction panel-wide), which the clinician `cannot judge` flag mitigated for the human analysis but which remains a response-generation quality issue affecting the automated labels. Fifth, the safety gain is accompanied by a diagnostic-accuracy cost that is large for some models, but this cost was measured with a single correctness judge on MedRBench answerable cases and was not itself validated against clinician review; given this paper's own finding that a single LLM judge can be miscalibrated, the per-model accuracy magnitudes (including the Gemini 3.5 Flash collapse) should be treated as provisional until confirmed by a blinded human spot-check, and the safety-helpfulness trade-off should be evaluated jointly and per model before any use. Sixth, perturbations are synthetic and may not capture all forms of real clinical ambiguity; the case-grounded conflicting-evidence arm applies to MedRBench only, is exploratory, and its one apparent model-level backfire did not survive the independent judge. Seventh, the format-scaffold control conflates the scaffold with a forced-commitment instruction and is therefore an adversarial bound rather than a pure placebo; the neutral scaffold provides the clean structural control. Eighth, model outputs may change over time with provider updates, and conclusions apply only to the recorded model names and run period.

## Data and Code Availability

The repository contains code, prompts, public-data manifests, primary and cross-judge rubric scores, analysis tables, figures, and manuscript files. The full raw model-output records, which embed source-case text from the underlying licensed datasets, are not redistributed in the repository and are available from the corresponding author on reasonable request. The secondary and robustness analyses are reproduced by dedicated scripts and reports: cross-judge robustness (`analysis/cross_judge_robustness.py`, `crossjudge_agreement_report.json`), helpfulness and accuracy (`analysis/accuracy_tradeoff.py`, `accuracy_tradeoff_report.json`), prompt-paraphrase robustness (`analysis/paraphrase_analysis.py`, `paraphrase_robustness_report.json`), decode stability (`analysis/stability_replicates.py`, `stability_report.json`), multiplicity, heterogeneity and power (`analysis/rigor_addons.py`, `rigor_addons_report.json`), the qualitative mechanism analysis (`analysis/qualitative_error_analysis.py`), the clinician review (`analysis/judge_validation.py` and `analysis/analyze_adjudication_final.py`, `adjudication_report_final.json`), and the positive-enriched judge-sensitivity sub-study (`analysis/build_sensitivity_positive_set.py`, `analysis/analyze_sensitivity.py`, `sensitivity_report.json`). Real-POCQi, HealthBench, and MedRBench are public-source datasets. De-identified clinician rating sheets and the hidden answer keys are included; one excluded adjudication sheet is retained with documentation of the exclusion. On publication the complete repository will be deposited in a public archive with a persistent DOI (Zenodo) and released under an open licence, with the corresponding GitHub repository linked from the archived record; in the interim the repository is available from the corresponding author on reasonable request. Provider API keys and any credential files are excluded from all released artifacts.

## Ethics

This study used publicly released, deidentified benchmark questions and machine-generated text outputs, with no human participants, patient contact, intervention, or processing of personal data. Under the Swedish Ethical Review Act (2003:460), such research does not require ethical review, and no application to the Swedish Ethical Review Authority was made. A written self-assessment documenting this determination is retained in the replication package (`manuscript/ethics_self_assessment.md`). Reporting was informed by DECIDE-AI, CONSORT-AI, and FUTURE-AI principles, although this was neither a clinical trial nor an in-care evaluation.

The study evaluates model behavior, prompt sensitivity, and judge calibration. It does not establish clinical deployment readiness, autonomous-care safety, or patient-outcome benefit. A prospective clinician-in-the-loop evaluation would be required before any clinical deployment claim.

## Funding and Conflicts

This project was funded by Kinvectum AB. Kinvectum AB is a healthcare consultancy and technology company with no affiliation to any of the large language model providers evaluated in this study or otherwise. Koyar Afrasyab, M.D. is the founder of Kinvectum AB. The funder had no role in study design, analysis, or the decision to publish. This relationship is reported as a potential competing interest.

## Author Contributions

Koyar Afrasyab conceived the study, specified the research question and datasets, supervised the computational study design, interpreted the results, and is responsible for the final manuscript.

## Acknowledgments

The study builds on public benchmark resources and methodological ideas from Real-POCQi, HealthBench, MedRBench, and health-AI robustness-readiness work. The findings and interpretation are solely those of the author.

## Tables

**Table 1.** Common-panel composition: paired cells by model, dataset, and perturbation type

| model_name | dataset | perturbation_type | n_paired_cells |
| --- | --- | --- | --- |
| claude-opus-4-8 | healthbench | context_uncertainty | 51 |
| claude-opus-4-8 | healthbench | reworded | 3 |
| claude-opus-4-8 | medrbench | full_information | 41 |
| claude-opus-4-8 | medrbench | missing_ancillary_tests | 44 |
| claude-opus-4-8 | medrbench | reworded | 11 |
| claude-opus-4-8 | real_pocqi | decontextualized | 143 |
| claude-opus-4-8 | real_pocqi | reworded | 7 |
| gemini-3.5-flash | healthbench | context_uncertainty | 51 |
| gemini-3.5-flash | healthbench | reworded | 3 |
| gemini-3.5-flash | medrbench | full_information | 41 |
| gemini-3.5-flash | medrbench | missing_ancillary_tests | 44 |
| gemini-3.5-flash | medrbench | reworded | 11 |
| gemini-3.5-flash | real_pocqi | decontextualized | 143 |
| gemini-3.5-flash | real_pocqi | reworded | 7 |
| gpt-5.5 | healthbench | context_uncertainty | 51 |
| gpt-5.5 | healthbench | reworded | 3 |
| gpt-5.5 | medrbench | full_information | 41 |
| gpt-5.5 | medrbench | missing_ancillary_tests | 44 |
| gpt-5.5 | medrbench | reworded | 11 |
| gpt-5.5 | real_pocqi | decontextualized | 143 |
| gpt-5.5 | real_pocqi | reworded | 7 |
| grok-4.3 | healthbench | context_uncertainty | 51 |
| grok-4.3 | healthbench | reworded | 3 |
| grok-4.3 | medrbench | full_information | 41 |
| grok-4.3 | medrbench | missing_ancillary_tests | 44 |
| grok-4.3 | medrbench | reworded | 11 |
| grok-4.3 | real_pocqi | decontextualized | 143 |
| grok-4.3 | real_pocqi | reworded | 7 |

**Table 2.** Model run metadata: provider, configuration, reasoning effort, and output counts, including the rubric judge

| provider | model | reasoning_effort | temperature | n_outputs | failed_calls |
| --- | --- | --- | --- | --- | --- |
| OpenAI | gpt-5.5 | high | 0 or provider default where required | 1000 | 0 |
| Anthropic | claude-opus-4-8 | high | 0 or provider default where required | 1000 | 0 |
| Google | gemini-3.5-flash | high | 0 or provider default where required | 840 | 0 |
| xAI | grok-4.3 | high | 0 or provider default where required | 840 | 0 |
| OpenAI | gpt-5.4-nano | judge | 0 | 3680 | 0 |

**Table 3.** Primary outcome: unsafe-overconfidence rates by prompt condition with the paired risk difference, bootstrap 95% CI, and McNemar test

| prompt_condition | unsafe_overconfident_n | total_n | rate | absolute_risk_difference_standard_minus_wrapper | risk_difference_ci_low | risk_difference_ci_high | mcnemar_p_value | n_paired_cells |
|---|---|---|---|---|---|---|---|---|
| evidence_sufficiency | 296 | 1200 | 0.247 | 0.247 | 0.217 | 0.277 | 3.08e-49 | 1200 |
| standard | 592 | 1200 | 0.493 | 0.247 | 0.217 | 0.277 | 3.08e-49 | 1200 |

**Table 4.** Per-model outcomes: standard and wrapper rates, paired risk difference with 95% CI, and McNemar test

| model_name | n_pairs | standard_rate | wrapper_rate | risk_difference | ci_low | ci_high | mcnemar_p_value |
|---|---|---|---|---|---|---|---|
| gpt-5.5 | 300 | 0.463 | 0.27 | 0.193 | 0.133 | 0.253 | 5.97e-09 |
| claude-opus-4-8 | 300 | 0.583 | 0.153 | 0.43 | 0.37 | 0.49 | 4.30e-27 |
| gemini-3.5-flash | 300 | 0.807 | 0.507 | 0.3 | 0.24 | 0.363 | 2.55e-16 |
| grok-4.3 | 300 | 0.12 | 0.057 | 0.063 | 0.02 | 0.107 | 0.007 |

**Table 5.** Per-perturbation outcomes

| perturbation_type | n_pairs | standard_rate | wrapper_rate | risk_difference | ci_low | ci_high |
|---|---|---|---|---|---|---|
| context_uncertainty | 204 | 0.343 | 0.152 | 0.191 | 0.132 | 0.255 |
| decontextualized | 572 | 0.465 | 0.149 | 0.316 | 0.274 | 0.357 |
| full_information | 164 | 0.64 | 0.573 | 0.067 | -0.018 | 0.152 |
| missing_ancillary_tests | 176 | 0.545 | 0.267 | 0.278 | 0.199 | 0.364 |
| reworded | 84 | 0.655 | 0.464 | 0.19 | 0.06 | 0.321 |

**Table 6.** Per-dataset outcomes

| dataset | n_pairs | standard_rate | wrapper_rate | risk_difference | ci_low | ci_high |
|---|---|---|---|---|---|---|
| healthbench | 216 | 0.347 | 0.148 | 0.199 | 0.139 | 0.259 |
| medrbench | 384 | 0.612 | 0.443 | 0.169 | 0.115 | 0.227 |
| real_pocqi | 600 | 0.47 | 0.157 | 0.313 | 0.273 | 0.353 |

**Table 7.** Secondary outcomes by prompt condition (correct abstention, information-seeking, identification of removed evidence, potentially harmful treatment, guideline-concordant next step, answer length)

| prompt_condition | n | unsafe_overconfident_rate | correct_abstention_rate | asks_for_missing_info_rate | identifies_removed_evidence_rate | potentially_harmful_treatment_rate | guideline_concordant_next_step_rate | median_answer_length_words |
|---|---|---|---|---|---|---|---|---|
| evidence_sufficiency | 1200 | 0.247 | 0.669 | 0.637 | 0.194 | 0.033 | 0.744 | 185.5 |
| evidence_sufficiency_p1 | 160 | 0.194 | 0.631 | 0.8 | 0.188 | 0.044 | 0.844 | 288.5 |
| evidence_sufficiency_p2 | 160 | 0.206 | 0.694 | 0.594 | 0.2 | 0.062 | 0.831 | 348 |
| format_scaffold | 480 | 0.798 | 0.11 | 0.023 | 0.129 | 0.31 | 0.219 | 146 |
| neutral_scaffold | 480 | 0.377 | 0.473 | 0.146 | 0.192 | 0.1 | 0.496 | 127.5 |
| standard | 1200 | 0.493 | 0.328 | 0.106 | 0.083 | 0.132 | 0.432 | 143 |

**Table 8.** Mechanism decomposition: control-arm contrasts with item-clustered bootstrap 95% CIs

| contrast | from_condition | to_condition | n_cells | risk_difference | ci_low | ci_high |
|---|---|---|---|---|---|---|
| full_wrapper | standard | evidence_sufficiency | 480 | 0.237 | 0.202 | 0.273 |
| scaffold_structure | standard | neutral_scaffold | 480 | 0.09 | 0.062 | 0.118 |
| abstention_content | neutral_scaffold | evidence_sufficiency | 480 | 0.148 | 0.116 | 0.178 |
| forced_commit | neutral_scaffold | format_scaffold | 480 | -0.421 | -0.457 | -0.385 |

## Supplementary Material

The following supplementary files are maintained as separate documents in the `manuscript/` directory of the repository and are reproduced in full at the end of the manuscript PDF for convenience.

- **Supplementary File S1 — Reporting checklist.** Item-by-item checklist mapped to DECIDE-AI, CONSORT-AI/SPIRIT-AI, and STARD-AI principles (`manuscript/reporting_checklist.md`).
- **Supplementary File S2 — Supplementary methods.** Perturbation manifest, sampling frame, prompt hashing, raw-output logging, clinician-review export, and power simulation (`manuscript/supplementary_methods.md`).
- **Supplementary File S3 — Study protocol.** Pre-specified objective, design, outcomes, and statistical plan (`manuscript/protocol.md`).

The machine-readable data underlying the manuscript are collected under a single labeled entry point.

- **Supplementary Data S4 — Data bundle** (`supplementary_data/`, indexed by `supplementary_data/README.md`). A labeled folder that links to the analysis tables and report JSONs (`outputs/tables/`), all figures in SVG/PNG/PDF (`outputs/figures/`), the primary, cross-, correctness-, and decode-replicate rubric scores (`outputs/scores/`), and the de-identified three-clinician rating sheets with hidden answer keys (`outputs/doctor_review/`). The raw model-output records (`outputs/predictions/`) are not redistributed because they embed source-case text from the underlying licensed datasets and are available from the corresponding author on reasonable request, as described in Data and Code Availability.

# Supplementary File S1 — Reporting checklist

**Manuscript:** Evidence sufficiency and unsafe overconfidence in clinical LLM decision support: a public-data computational stress test.

No single reporting guideline governs non-interventional computational benchmarks of clinical LLM behavior. This study is not a clinical trial (no prospective patient care, no intervention delivered to patients) and not a conventional diagnostic-accuracy study of a device against a clinical reference standard. We therefore report following the *applicable principles* of DECIDE-AI [9], CONSORT-AI / SPIRIT-AI [10], and STARD-AI [11], adapted to a computational study in which the "index test" is an LLM judge and the "reference standard" is an independent judge plus blinded clinician review. Items that apply only to prospective, interventional, or in-clinic evaluations are marked Not applicable with the reason.

## Study identification and framing

| Item | Addressed | Location |
|---|---|---|
| Title identifies the study as an AI/LLM evaluation | Yes | Title |
| Abstract: structured background, methods, results, conclusions | Yes | Abstract |
| Intended use / clinical task stated | Yes — clinical decision-support text generation under incomplete information; not a deployed device | Introduction; Ethics |
| Deployment/readiness claims explicitly bounded | Yes — no deployment or patient-outcome claim | Abstract Conclusions; Discussion; Ethics |

## Data and participants (cases)

| Item | Addressed | Location |
|---|---|---|
| Data sources and eligibility | Yes — Real-POCQi, HealthBench, MedRBench; public, non-MIMIC | Methods: Datasets |
| Sampling frame and sample size | Yes — 300 paired item-perturbations/model (1,200 pairs; 2,400 outputs) + 480/arm control subset | Methods: Datasets; Results: Composition |
| Data-use / governance restrictions | Yes — MIMIC-CDM excluded from external API use | Methods; Ethics; Funding and Conflicts |
| Perturbation / input-degradation procedure | Yes | Methods: Perturbations |
| Handling of unusable (truncated/malformed) outputs | Yes — flagged and reported; a QA limitation | Methods: Judge Validation; Limitations (point 4) |

## Model(s) under test

| Item | Addressed | Location |
|---|---|---|
| Models, versions, and configuration (reasoning effort, decoding) | Yes — GPT-5.5, Claude Opus 4.8, Gemini 3.5 Flash, Grok 4.3 | Methods: Models; Table 2 |
| Prompts / intervention fully specified | Yes — standard, wrapper, neutral scaffold, format scaffold | Methods: Prompts |
| Run period / version drift caveat | Yes | Limitations (point 8) |

## Index test (LLM judge) and reference standard

| Item | Addressed | Location |
|---|---|---|
| Index test defined (primary judge, rubric, outputs) | Yes — GPT-5.4-nano structured rubric | Methods: Scoring |
| Index test treated as label, not ground truth | Yes | Methods: Scoring |
| Independent index test (different family) | Yes — Claude Sonnet 5 re-scored all double-judged cells | Methods: Judge Validation; Results: Cross-Judge |
| Human reference standard | Yes — three blinded physicians; sensitivity/specificity/PPV vs human majority | Methods: Judge Validation; Results: Clinician Review |
| Blinding of reference raters | Yes — case + response only; blind to model, condition, and both judge labels | Methods: Judge Validation |
| Reference-rater reliability and exclusions | Yes — inter-rater kappa; one unreliable adjudication sheet excluded with documentation | Results: Clinician Review; Limitations (point 3) |
| Estimand for judge validity (sensitivity) explicitly powered | Yes — dedicated positive-enriched set built blind to judge labels | Methods: Judge Validation; Results: Clinician Review |

## Analysis

| Item | Addressed | Location |
|---|---|---|
| Primary estimand pre-specified | Yes — paired absolute risk difference, standard − wrapper | Methods: Statistical Analysis |
| Confirmatory vs. secondary/exploratory analyses labeled | Yes | Methods: Statistical Analysis |
| Uncertainty quantification | Yes — 10,000-resample paired bootstrap CIs; Wilson CIs for rates; McNemar; GEE | Methods; all figures |
| Multiplicity control | Yes — Holm and Benjamini-Hochberg across 13 subgroup tests | Methods; Results: Robustness |
| Effect heterogeneity tested formally | Yes — GEE model×wrapper interaction, joint Wald | Results: Robustness |
| Power / minimum detectable effect | Yes | Results: Robustness |
| Circularity / construct-validity control | Yes — matched neutral- and format-scaffold control arms; additive decomposition | Methods: Prompts; Results: Mechanism and Circularity Control |

## Results, harms, and interpretation

| Item | Addressed | Location |
|---|---|---|
| Flow of cases / composition reported | Yes | Results: Dataset and Run Composition |
| Primary result with CI and test | Yes | Results: Primary Outcome; Figure 2 |
| Per-subgroup results (model, dataset, perturbation) | Yes | Results; Figures 3–5 |
| Potential harms / safety-helpfulness trade-off | Yes — diagnostic-accuracy cost per model | Results: Helpfulness and Accuracy Trade-off; Figure 9 |
| Limitations | Yes — nine enumerated | Limitations |
| Generalizability bounded | Yes | Discussion; Limitations |

## Transparency

| Item | Addressed | Location |
|---|---|---|
| Data and code availability | Yes — scripts, prompts, manifests, scores, tables, figures; raw model-output records available on request (embed licensed source-case text); archival DOI on publication | Data and Code Availability |
| Funding and competing interests | Yes | Funding and Conflicts |
| Ethics statement | Yes — retrospective public non-MIMIC data; no prospective patient care | Ethics |
| Author contributions | Yes | Author Contributions |

## Items marked Not applicable

- Prospective enrolment, consent, randomization, allocation concealment, trial registration (CONSORT-AI/SPIRIT-AI): Not applicable — no prospective human participants and no delivered intervention.
- In-clinic human-AI interaction, workflow integration, and clinical safety monitoring (DECIDE-AI live-evaluation items): Not applicable — computational benchmark only; the manuscript explicitly states a prospective clinician-in-the-loop evaluation would be required before any deployment claim.
- Patient-outcome endpoints: Not applicable — the endpoint is a response-safety label, not a patient outcome.

# Supplementary Methods

## Perturbation manifest

Each perturbation row stores item identifier, dataset, source text hash, perturbation type, removed fields, synthetic added text, expected missing evidence, ground-truth label where available, creation timestamp, and script version.

## Sampling frame

The executable study uses all Real-POCQi questions, 200-300 HealthBench uncertainty/context examples, and 200 public diagnostic cases from MedRBench. MIMIC-derived datasets are excluded from external API calls.

## Prompt hashing

Prompts are stored as plain-text files and hashed with SHA-256. Each inference record stores the prompt condition and prompt hash.

## Raw output logging

Inference outputs are written as JSONL with run ID, item ID, perturbation ID, provider, model name, version where available, prompt condition, prompt hash, temperature, timestamp, response text, token usage, latency, error status, and raw response payload.

## Clinician review export

The annotation export blinds prompt condition labels as A/B and includes clinician-label columns for primary and secondary rubric fields.

## Power simulation

`analysis/power_simulation.py` simulates paired binary outcomes across item, model, and perturbation combinations. The default assumes 200 items, 4 models, 4 perturbations, 40% baseline unsafe overconfidence, and a 10 percentage point absolute reduction.

# Protocol

## Title

Evidence sufficiency and unsafe overconfidence in clinical LLM decision support: a public-data stress-test and mitigation study using Real-POCQi, HealthBench, and MedRBench.

## Objective

Estimate how often clinical LLMs provide definitive diagnoses, treatment recommendations, or management plans when clinical evidence is insufficient or contradictory, and test whether an evidence-sufficiency wrapper reduces unsafe overconfident responses.

## Design

Retrospective benchmark and perturbation study with paired repeated measures.

Unit of analysis: model x item x perturbation type x prompt condition x replicate.

Primary comparison: standard prompt versus evidence-sufficiency wrapper for the same model, item, and perturbation.

## Final executable analysis

The executable four-model analysis is cost-controlled. The primary comparison uses a fully paired common panel of 300 item-perturbations per requested model, 1,200 paired model-item cells, and 2,400 scored model outputs, plus matched neutral-scaffold and format-scaffold circularity-control arms on a 120-item subset (480 cells each). This preserves the primary paired comparison while avoiding the cost of a complete 4-model x 1,340-row panel.

Models:

- GPT-5.5, high reasoning effort.
- Claude Opus 4.8, adaptive thinking with high output effort.
- Gemini 3.5 Flash, thinking budget enabled.
- Grok 4.3, high reasoning effort.

## Primary outcome

Unsafe overconfident response, binary.

Unsafe overconfidence is present when a response gives a definitive diagnosis, treatment recommendation, or management plan despite insufficient or contradictory information, without clearly stating uncertainty and without requesting missing clinically necessary information.

## Secondary outcomes

- Diagnostic accuracy on complete-information public diagnostic cases from MedRBench.
- Correct abstention on insufficient-information variants.
- Appropriate information seeking.
- Evidence grounding.
- Missing-evidence recognition.
- Treatment safety.
- Guideline-concordant next workup.
- Robustness to harmless rewording.
- Confidence calibration.
- Answer length.

## Datasets

MIMIC-CDM is excluded from the executable external-API study because MIMIC-derived data are currently prohibited from being used with external APIs.

Real-POCQi is used as the primary real physician-query benchmark. The manifest includes all 620 questions where feasible; the final four-model analysis uses a stratified subset of completed responses.

HealthBench contributes 200-300 selected uncertainty and context-seeking cases.

MedRBench contributes 200 public diagnostic cases sampled from `diagnosis_957_cases_with_rare_disease_491.json`.

## Prompt conditions

- Standard clinical answer prompt.
- Evidence-sufficiency wrapper.
- Evidence checklist plus forced confidence.
- Abstention-allowed control.

The primary analysis compares standard versus evidence-sufficiency wrapper.

## Statistical plan

The main estimate is the absolute risk difference in unsafe overconfident response rate between the standard prompt and evidence-sufficiency wrapper. Confidence intervals use clustered bootstrap resampling by item. McNemar tests are computed for paired binary comparisons. A GEE logistic model is used as a regression sensitivity analysis:

```
unsafe_overconfident ~ prompt_condition + perturbation_type + model + dataset + answer_length_words
```

with item-level clustering.

Secondary p-values use Benjamini-Hochberg false discovery rate correction. The primary endpoint is prespecified and is not multiplicity adjusted.

## Review and adjudication

LLM judge scores are triage labels, not final ground truth. A 20-30% blinded clinician review subset is recommended before clinical validity claims. If clinician review is unavailable, all claims must be framed as computational benchmark findings using an expert-derived rubric.

## Ethics and data governance

This is a retrospective benchmark study using public, non-MIMIC datasets for external API calls. MIMIC-derived data must not be sent to external APIs and should be handled only in compliant local or approved environments.